\documentclass{article} 
\usepackage{iclr2027_conference,times}

\usepackage{amsmath,amsfonts,bm}

\def\eqref#1{equation~\ref{#1}}

\def\1{\bm{1}}

\DeclareMathAlphabet{\mathsfit}{\encodingdefault}{\sfdefault}{m}{sl}
\SetMathAlphabet{\mathsfit}{bold}{\encodingdefault}{\sfdefault}{bx}{n}

\usepackage{hyperref}
\usepackage{url}
\PassOptionsToPackage{table}{xcolor}

\usepackage[utf8]{inputenc}
\usepackage[T1]{fontenc}
\usepackage{microtype}
\usepackage{graphicx}
\usepackage{subcaption}
\usepackage{booktabs}
\usepackage{multirow}
\usepackage{colortbl}
\usepackage{tabularx}
\usepackage{enumitem}
\usepackage{amsmath}
\usepackage{amsfonts}
\usepackage{amssymb}
\usepackage{mathtools}
\usepackage{amsthm}
\usepackage{algorithm}
\usepackage{algorithmic}
\usepackage[capitalize,noabbrev]{cleveref}
\usepackage[table]{xcolor}
\usepackage{wrapfig}

\usepackage{enumitem}

\setlist[itemize]{nosep}
\setlist[enumerate]{nosep}
\usepackage{wrapfig}

\usepackage{enumitem}
\setlist[itemize]{nosep}
\setlist[enumerate]{nosep}
\theoremstyle{plain}

\theoremstyle{definition}

\theoremstyle{remark}

\title{DSDyn-VLA: A Dual-Stream Dynamic Manipulation Framework with Motion Perception, Future Awareness, and Realtime Correction}

\author{
Wenhao Li \\
University of Sydney
\And
Xiu Su \\
Central South University
\AND
Yu Han \\
University of California, San Diego
\And
Yichao Cao \\
Central South University
\AND
Shan You \\
Ace Robotics
\And
Chang Xu \\
University of Sydney
}

\iclrfinalcopy 
\begin{document}

\maketitle
\lhead{Preprint}

\begin{abstract}
While Vision-Language-Action (VLA) models excel in static tasks, they struggle in dynamic environments where objects are in motion (e.g., conveyor belt manipulation). We identify three fundamental limitations hindering current VLAs in these scenarios: the \textbf{perception gap}, where static visual inputs lack temporal motion cues; the \textbf{latency gap}, where inference delays render actions obsolete; and the \textbf{control gap}, caused by the open-loop action chunk execution without real-time adjustment. In this work, we propose \textbf{DSDyn-VLA}, a Slow-Fast \textbf{D}ual-\textbf{S}tream \textbf{Dyn}amic manipulation framework that integrates motion-aware foresighted planning with real-time residual correction. The slow \textbf{Flow-Planner} serves as a macro-planner. By  enhancing the VLA with optical flow for temporal perception and a future state awareness mechanism to preemptively offset inference latency, it produces globally consistent, motion-aware action chunks. Complementing this, the fast \textbf{Res-Refiner} employs a lightweight RL policy to inject high-frequency, closed-loop corrections into the planned action chunks based on real-time observations. In addition, we introduce \textbf{DynBench}, a MuJoCo-based benchmark for dynamic object manipulation that comprises nine tasks. Extensive experiments demonstrate that DSDyn-VLA reduces the failure rate by over 76\% compared to current SOTA method in high-latency setting on the Kinetix dynamic benchmark, while achieving about 6$\times$ the success rate of PI0.5 in real-world dynamic settings and about 5$\times$ on DynBench. We will open-source all the code and weights.
\end{abstract}

\section{Introduction}

In recent years, the field of robotic manipulation \cite{b1,b2,b3,b4,b5} has witnessed a paradigm shift with the emergence of Vision-Language-Action (VLA) models \cite{b6,b7,b8,b9,b10,b11}. By training on internet-scale multimodal data \cite{b15,b16}, these models have demonstrated remarkable capabilities in semantic reasoning and zero-shot generalization, enabling robots to interpret complex natural language instructions and perform diverse manipulation tasks \cite{b12,b13,b14}. Current VLAs have achieved high success rates across a broad spectrum of tasks, ranging from simple pick-and-place operations to complex activities such as folding clothes, assembling boxes, and making coffee \cite{b17,b18,b19}.

  \begin{figure*}[t]
    \centering
    \includegraphics[width=\textwidth]{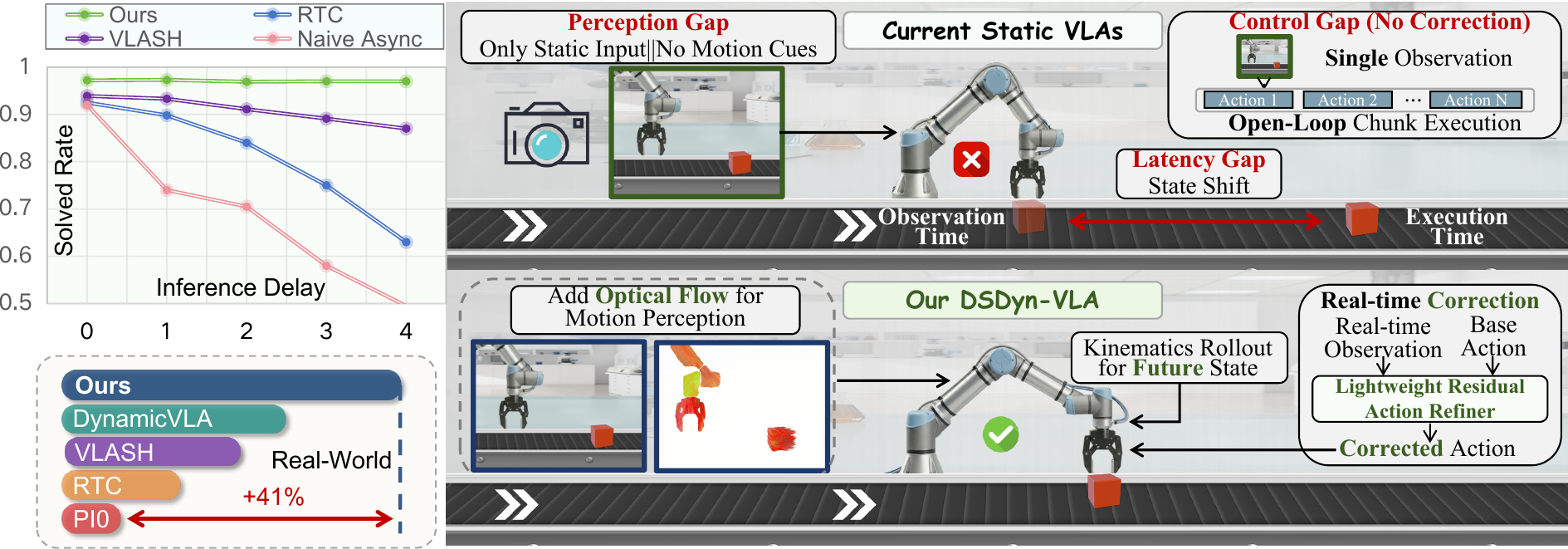}
    \caption{\textbf{Left:} Comparison on Kinetix benchmark and real-world tasks. \textbf{Top Right:} Current VLAs fail to perform dynamic tasks due to three gaps: the perception gap, latency gap, and control gap. \textbf{Bottom Right:} Our DSDyn-VLA addresses these gaps by introducing optical flow for motion perception, future state awareness for latency compensation, and a refiner for real-time correction.}
    \label{fig0}
    \vspace{-18pt}
  \end{figure*}

However, regardless of their complexity, all of these tasks are static, where target objects do not move spontaneously. Yet, the physical world is inherently dynamic. Many practical applications---ranging from industrial sorting on moving conveyor belts \cite{b20} to household human-robot handover \cite{b21} or intercepting moving objects \cite{b22}---require the robot not only to understand ``what to do'' but also to react with precise timing and agility. In these dynamic scenarios, the capabilities of current VLA models break down. We argue that the inability of existing VLAs to handle dynamic manipulation stems from three fundamental mismatches, which we categorize as the \textit{1) Perception Gap}, the \textit{2) Latency Gap}, and the \textit{3) Control Gap}.

The \textit{Perception Gap} arises because standard VLAs typically rely on static RGB snapshots as visual input \cite{b23}, which preserve semantics but discard temporal motion cues. Without explicit velocity-related information, a VLA cannot estimate the target object's trajectory or time-to-contact, which is a prerequisite for dynamic interaction. The \textit{Latency Gap} comes from the inference delay between observation and execution \cite{b24,b25}. While this delay is negligible in static settings, the world state evolves significantly during inference in dynamic environments. Consequently, actions based on obsolete observations inevitably lead to execution failures. The \textit{Control Gap} is caused by ``action chunking'' \cite{b26}. To ensure temporal consistency, VLAs predict a multi-step action chunk at once. This chunk is executed in an open-loop manner. Once chunk execution begins, the robot is ``blind'' to new environmental changes until the next inference cycle. This lack of closed-loop correction makes it impossible to adjust to perturbations or unexpected object movements within a chunk's duration.

To bridge these gaps, we propose \textbf{DSDyn-VLA}, a novel Slow-Fast Dual-Stream Framework inspired by the complementary roles of the cerebrum and cerebellum in motor control. The ``Slow'' stream, designated as the \textbf{Flow-Planner}, serves as the semantic macro-planner. It is an enhanced VLA model designed to overcome perception and latency gaps. To address the Perception Gap, the Flow-Planner augments the visual input with optical flow, explicitly providing the model with temporal dynamics and motion cues. For the Latency Gap, we introduce a Future State Awareness mechanism. Instead of planning for the \textit{current} state, the Flow-Planner predicts the future state at execution time and generate actions for that future moment, thereby preemptively offsetting system latency.

Complementing this, the ``Fast'' stream, termed the \textbf{Res-Refiner}, acts as a reactive motion refiner. Res-Refiner consists of a lightweight, RL-trained residual network. It operates in real-time, running in parallel with the execution of the Flow-Planner's action chunks. By taking real-time observations, the Res-Refiner injects residual corrections into the base action each timestep. This effectively closes the control loop, allowing the system to compensate for prediction errors and rapid environmental changes that the Flow-Planner cannot react to in time, thus resolving the Control Gap. Extensive experiments demonstrate that DSDyn-VLA reduces the failure rate of SOTA methods from 13\% to 3\% on the Kinetix dynamic benchmark, while improving the real-world success rate from 8.4\% with PI0.5 to 49.0\%. In addition, we construct \textbf{DynBench}, a MuJoCo-based dynamic manipulation simulation benchmark, comprising 9 tasks, on which DSDyn-VLA achieves an average success rate of 48.4\%, substantially outperforming PI0.5 at 9.7\%. Our main contributions are as follows:
\begin{itemize}
\item We propose \textbf{DSDyn-VLA}, a novel Slow-Fast Dual-Stream framework tailored for dynamic manipulation. It synergizes macro-level chunk prediction with stepwise residual correction to resolve the real-time responsiveness limitations of traditional VLAs.
\item We design the \textbf{Flow-Planner}, an enhanced VLA slow stream that integrates optical flow input with a future state awareness mechanism, endowing the VLA with the capability to perceive object motions and proactively offset inference latency.
\item We introduce the \textbf{Res-Refiner}, a real-time RL-based residual policy that performs high-frequency closed-loop adjustments during action chunk execution, effectively compensating for environmental perturbations to ensure robust performance in highly dynamic settings.
\item We construct \textbf{DynBench}, which establish a new testbed for systematically evaluating dynamic manipulation policies.
\end{itemize}

\section{Related Work}


\paragraph{Real-time VLA Models for Latency Gap.} To mitigate the ``stop-and-go'' behavior caused by inference latency, recent works have adopted asynchronous inference \cite{b27,b28}. However, this introduces temporal misalignment, as observations become stale during generation. Real-Time Chunking (RTC) \cite{b24} addresses this via inference-time inpainting, while Training-Time RTC \cite{b29} optimizes this by conditioning on action prefixes during training to eliminate runtime overhead. Similarly, VLASH \cite{b25} bridges the gap by rolling forward the robot's proprioceptive state to condition generation on estimated execution-time states, which inspires the state prediction mechanism in our Flow-Planner. DynamicVLA \cite{b38} mitigates the latency gap through latency-aware action streaming, which discards stale actions. However, these methods still rely on static visual inputs and open-loop chunk execution. Consequently, they fail to address the \textit{Perception Gap} and \textit{Control Gap}, causing them to remain ineffective for highly dynamic tasks.



\section{Method}
\label{sec:method}

\subsection{Preliminaries}
\label{sec:prelim}
\textbf{Optical Flow for Dynamics Perception.} Standard VLA models operate on static RGB snapshots $\mathbf{I}_{rgb} \in \mathbb{R}^{H \times W \times 3}$. However, this representation lacks the high-order derivative information---specifically velocity---which is crucial for moving targets. Optical flow field $\mathbf{F}_t \in \mathbb{R}^{H \times W \times 2}$ is a dense vector map where each pixel $(x, y)$ is assigned a displacement vector $\mathbf{v}_{x,y} = (u, v)$, representing the motion state between consecutive frames $\mathbf{I}_{t-1}$ and $\mathbf{I}_t$. Unlike raw pixel differences, $\mathbf{F}_t$ explicitly encodes the scene's motion field. By conditioning the policy on $\mathbf{F}_t$, we effectively provide the model with a visual approximation of the system's state derivative.

\begin{wrapfigure}{r}{0.48\columnwidth}
  \vspace{-8pt}
  \centering
  \includegraphics[width=\linewidth]{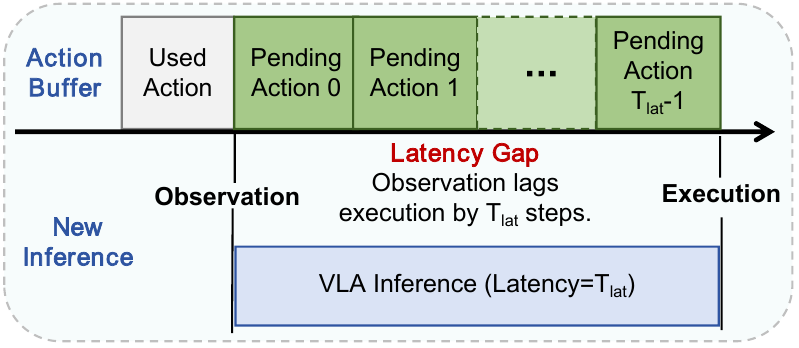}
  \vspace{-15pt}
  \caption{Illustration of VLA asynchronous inference. VLA initiates a new inference when $T_{lat}$ (representing the VLA inference latency steps) actions remain in the action buffer, ensuring the new action chunk is available when the buffer is exhausted. The observation for inference lags the  execution by $T_{lat}$ steps.}
  \label{fig2}
  \vspace{-5pt}
\end{wrapfigure}

\textbf{Action Chunking Policy.}To ensure temporal coherence, most existing VLA models adopt an Action Chunking mechanism. At time step $t$, the policy $\pi$ maps the current observation $\mathbf{o}_t$ to a sequence of $H$ future actions:
\begin{equation}
    \mathbf{A}_t = \pi(\mathbf{o}_t) = \{ \mathbf{a}_k \}_{k=0}^{H-1}
\end{equation}
The robot executes a subset of these actions (execution horizon $K \le H$) before the next inference cycle completes. This chunk-based generation is essential for high-frequency control but inherently transitions the system from fully closed-loop feedback to a partially open-loop execution. During the horizon $K$, the robot executes the pre-planned sequence blindly, rendering it unresponsive to real-time environmental perturbations.

\textbf{Asynchronous Inference and Latency Gap.} Deploying VLA models for real-time control typically adopts an Asynchronous Inference paradigm. As illustrated in Figure \ref{fig2}, the inference process is decoupled from robot execution. To ensure continuous motion, the VLA initiates a new prediction while the previous action chunk is still being executed, ensuring that the fresh action chunk arrives before the current buffer is exhausted. Let $T_{lat}$ denote the inference latency in steps; the VLA triggers inference $T_{lat}$ steps in advance. This delay creates a fundamental \textit{State Shift}, which we term the \textit{Latency Gap}. The policy generates actions conditioned on the past state $\mathbf{s}_{t}$, but these actions are applied to the future state $\mathbf{s}_{t+T_{lat}}$. In dynamic settings where the environment evolves significantly during $T_{lat}$, this temporal misalignment leads to catastrophic errors.

\subsection{DSDyn-VLA Overview}
\label{sec:overview}

\begin{figure*}[t!]
  \centering
\includegraphics[width=\textwidth]{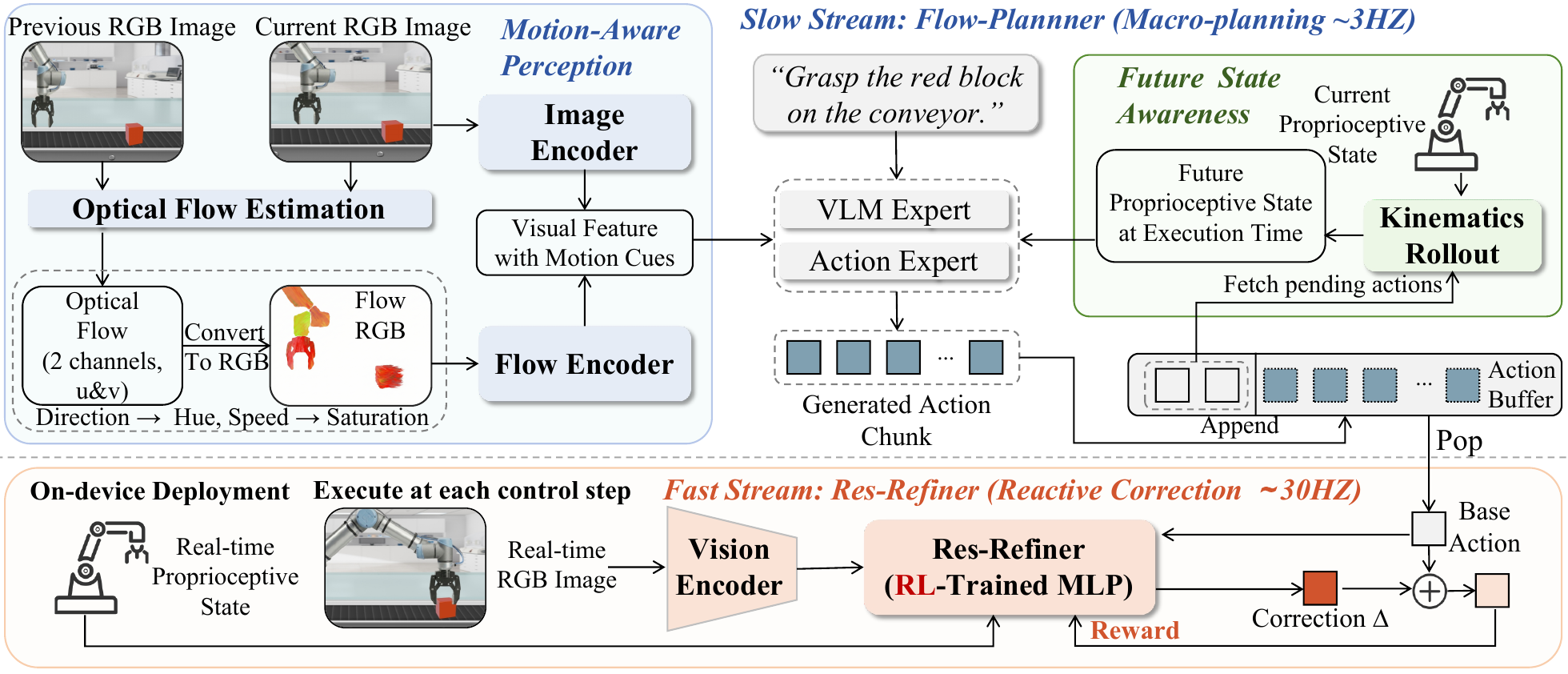}
\vspace{-15pt}
\caption{Architecture of our DSDyn-VLA, which consists of Flow-Planner and Res-Refiner.}
  \label{fig1}
  \vspace{-15pt}
\end{figure*}

We propose \textbf{DSDyn-VLA}, a bi-level framework inspired by the biological interplay between the cerebrum and cerebellum. The system decomposes the control problem into two parallel streams operating at distinct frequencies:

\begin{itemize}
    \item \textbf{The Slow Stream (Flow-Planner):} Operates at a lower frequency ($f_{slow} \approx 2-3$Hz). It utilizes a high-capacity VLA backbone to perform semantic reasoning and long-horizon planning. It addresses the \textit{Perception Gap} via optical flow integration and the \textit{Latency Gap} via a future-state awareness mechanism.
    \item \textbf{The Fast Stream (Res-Refiner):} Operates at a high frequency ($f_{fast} \approx 30$Hz). It utilizes a lightweight, edge-deployed residual policy to perform real-time closed-loop corrections. It addresses the \textit{Control Gap} by reacting to high-frequency disturbances that occur within the duration of a macro-action chunk.
\end{itemize}

Formally, the final control action $\mathbf{a}_t^{final}$ executed by the robot is the superposition of the open-loop macro-plan and the closed-loop residual:
\begin{equation}
    \mathbf{a}_t^{final} = \underbrace{\mathbf{a}_{t}^{slow}}_{\text{Macro-Plan}} + \underbrace{\pi_{fast}(\mathbf{s}_t, \mathbf{a}_{t}^{slow})}_{\text{Reflex-Correction}}
\end{equation}
where $\mathbf{a}_{t}^{slow}$ is the specific step corresponding to time $t$ extracted from the active chunk $\mathbf{A}_{slow}$.

\subsection{The Slow Stream: Flow-Planner}
\label{sec:slow_stream}

The Flow-Planner serves as the semantic backbone of our system. It is built upon the PI0.5 architecture \cite{b14}, a Conditional Flow Matching (CFM) policy. We introduce substantial architectural enhancements to adapt it for dynamic interaction.

\subsubsection{Motion-Aware Perception}
To inject explicit temporal cues into the VLA latent space, we design a dual-branch visual system.

\textbf{Flow Extraction \& Visualization:} First, we extract the dense optical flow $\mathbf{F}_t$ using RAFT-Large \cite{b34}, a recurrent all-pairs field transform network. To align this geometric data with the pre-trained semantic space of CLIP, we transform the 2-channel flow field into a 3-channel RGB-like representation $\mathbf{I}_{flow}$ via a differentiable polar mapping. For a flow vector $(u, v)$:
\begin{equation}
    \text{Hue} = \text{atan2}(v, u), \quad \text{Saturation} = \sqrt{u^2 + v^2}
\end{equation}
This mapping visualizes motion direction as color and speed as intensity.

\textbf{Dual-Encoder Architecture:} We employ two parallel Vision Transformers as encoders. The first encoder $E_{rgb}$ processes the static image $\mathbf{I}_{rgb}$ to extract semantic features (e.g., object identity). The second encoder $E_{flow}$ processes the flow image $\mathbf{I}_{flow}$ to extract dynamic features (e.g., motion trajectory).
We fine-tune both encoders end-to-end. The output tokens are concatenated along the sequence dimension to form the unified visual context:
\begin{equation}
    \mathbf{z}_{vis} = \text{Concat}\left( E_{rgb}(\mathbf{I}_{rgb}), E_{flow}(\mathbf{I}_{flow}) \right) \in \mathbb{R}^{(2N) \times D}
\end{equation}
where $N$ is the number of patches and $D$ is the embedding dimension. This ensures the planner attends to both ``what'' the object is and ``how'' it is moving.

\subsubsection{Future State Awareness via Kinematics Rollout}
To neutralize the \textit{Latency Gap} defined in Sec. \ref{sec:prelim}, the Flow-Planner must condition its generation on the robot's state at the \textit{execution time}, rather than the \textit{observation time}. Since the visual future is unknown, we exploit the deterministic nature of the robot's internal state.

We propose a \textit{Kinematics Rollout} mechanism. During the inference interval $T_{lat}$, the robot continues to execute actions from the previous chunk's buffer $\mathbf{A}_{buffer}$. We can explicitly calculate the future proprioceptive state $\hat{\mathbf{q}}_{future}$ by simulating the forward kinematics $f_{kin}$:
\begin{equation}
    \hat{\mathbf{q}}_{future} = f_{kin}\left( \mathbf{q}_{query}, \sum_{i=0}^{T_{lat}} \mathbf{A}_{buffer}[i] \right)
\end{equation}
This predicted state $\hat{\mathbf{q}}_{future}$ is  injected into the VLA as the proprioceptive conditioning. To align the model with this inference logic, we employ a randomized time-shift strategy during training. By pairing the visual observation at $t$ with the proprioceptive state and action targets from a stochastically sampled future timestep $t+\delta$, the model is forced to correlate past visual cues with future proprioceptive states across varying temporal scales. This ensures the planner remains robust to fluctuating system latencies. Consequently, the generative model learns the mapping:
\begin{equation}
    (\text{Past Vision}, \text{Future Robot State}) \xrightarrow{\pi} \text{Future Actions}
\end{equation}
This effectively shifts the planner's temporal frame of reference to the moment of execution, eliminating the state drift caused by latency.


\subsection{The Fast Stream: Res-Refiner}
\label{sec:fast_stream}

While the Flow-Planner ensures global optimality and handles latency, it operates open-loop within the execution horizon (typically 0.5s - 1s). The Res-Refiner bridges the \textit{Control Gap} by providing high-frequency (30Hz), closed-loop feedback.

The Res-Refiner adopts a pre-trained visual encoder and a multi-head actor-critic architecture optimized for pixel-based reinforcement learning.

\textbf{Spatial-Learned Visual Encoding:}
We employ a ResNet-10 backbone \cite{b35} followed by a Learnable Spatial Embedding layer. This layer uses a set of $N_s$ learned spatial blocks ($N_s=8$ in our implementation) to extract critical geometric features from the image feature map $\Phi \in \mathbb{R}^{C \times H' \times W'}$. The resulting representation $\mathbf{z}_{vis\_res} \in \mathbb{R}^{256}$ captures the precise spatial relationships between the robot end-effector and the object.

\textbf{Multi-Modal Policy Network:}
The policy network is a MLP with two hidden layers of 256 units each. It performs early fusion by concatenating the visual encoding $\mathbf{z}_{vis\_res}$ from multiple camera streams, the robot's proprioceptive state $\mathbf{q}_t$, and the macro-action $\mathbf{a}_t^{slow}$ provided by the planner. To prevent over-fitting and enhance robustness, we incorporate DrQ-style data augmentation, applying random pixel shifts ( $\pm 4$ pixels) to the visual input during training. The network outputs the residual correction $\Delta \mathbf{a}_t$ through a Tanh-squashed multivariate Normal distribution:
\begin{equation}
\begin{aligned}
    \mathbf{h} &= \text{Concat}(\mathbf{z}_{vis\_res}, \mathbf{q}_t, \mathbf{a}_t^{slow}) \\
    \Delta \mathbf{a}_t &= \alpha_{scale} \cdot \text{Tanh}\left( \text{MLP}(\mathbf{h}) \right)
\end{aligned}
\end{equation}

\subsection{Training Strategy}
\label{sec:training}

\textbf{Stage 1: Motion-Aware SFT with Time-Shift.} We first train the Flow-Planner using a dataset of dynamic demonstrations $\mathcal{D} = \{(\mathbf{I}_t, \mathbf{q}_t, \mathbf{A}_t)\}$. The objective is to minimize the Conditional Flow Matching loss.
To endow the model with Future State Awareness, we apply a Time-Shift Augmentation to the dataloader. For a training sample recorded at time $t$, we construct the input-target pair as follows:
 \textbf{Input:} Visual observation at time $t$ ($\mathbf{I}_t$) + Proprioceptive state at time $t+\Delta_{steps}$ ($\mathbf{q}_{t+\Delta}$).
 \textbf{Target:} Action chunk starting at time $t+\Delta_{steps}$ ($\mathbf{A}_{t+\Delta}$).
The loss function regresses the vector field $v_\theta$ against the target vector field $u_t$:
\begin{equation}
\begin{split}
    \mathcal{L}_{CFM}(\theta) = \mathbb{E}_{t, \mathbf{x}_0, \mathbf{A}_{gt}} \Big[ \| v_\theta(\psi_t(\mathbf{x}_0), t, \mathbf{z}_{vis}, \mathbf{q}_{t+\Delta}) \\
    - u_t(\psi_t(\mathbf{x}_0) | \mathbf{A}_{t+\Delta}) \|^2 \Big]
\end{split}
\end{equation}
This forces the model to exploit the motion dynamics encoded in $\mathbf{z}_{vis}$ to learn the causal relationship between current visual cues and future actions, implicitly modeling the latency physics.

\textbf{Stage 2: Residual RL Fine-Tuning.} In the second stage, we freeze the Flow-Planner parameters $\theta$ to preserve its semantic generalization. We train the Res-Refiner parameters $\phi$ using RLPD \cite{b36}, a sample-efficient RL algorithm.
The Res-Refiner interacts with the environment, receiving sparse binary rewards $r_{success}$. To accelerate learning, we maintain a Replay Buffer that mixes 50\% offline demonstrations and 50\% online experiences.
We utilize an ensemble of $N_c$ critics $\{Q_i\}_{i=1}^N$ to mitigate Q-value overestimation. The critic loss minimizes the Bellman error averaged over the ensemble:
\begin{equation}
\begin{split}
    \mathcal{L}_{critic}(\phi) = \frac{1}{N} \sum_{i=1}^N \mathbb{E}_{\mathcal{B}} \Big[ \Big( Q_i(\mathbf{s}, \mathbf{a}) \\
    - (r + \gamma \max_{\mathbf{a}'} \min_{j} Q_j(\mathbf{s}', \mathbf{a}')) \Big)^2 \Big]
\end{split}
\end{equation}
The actor is updated to maximize the Q-value subject to entropy regularization. Through this stage, the Res-Refiner learns to compensate for the noise and dynamic perturbations that the open-loop Flow-Planner cannot anticipate.

\begin{table*}[t!]
\centering
\footnotesize
\vspace{-15pt}
\caption{Real-world success rates on dynamic manipulation tasks (5 repeated 50-trials).}
\vspace{-5pt}
\label{tab:real_world_results}
\begin{tabular*}{\textwidth}{@{\extracolsep{\fill}}lccccc}
\toprule
Task & Baseline (PI0.5) & RTC & VLASH & DynamicVLA & \textbf{DSDyn-VLA (Ours)} \\
\midrule
Conveyor Picking & 8.8$\pm$3.6\% & 12.4$\pm$3.8\% & 20.0$\pm$4.5\% & 33.2$\pm$5.4\% & \textbf{44.8$\pm$4.1\%} \\
Dynamic Dropping & 14.8$\pm$4.6\% & 43.6$\pm$5.7\% & 50.0$\pm$5.5\% & 56.0$\pm$5.1\% & \textbf{62.8$\pm$5.4\%} \\
Dynamic Stacking & 1.2$\pm$1.8\% & 2.4$\pm$2.6\% & 6.8$\pm$4.1\% & 18.0$\pm$5.1\% & \textbf{32.0$\pm$4.5\%} \\
Mobile Pouring & 8.8$\pm$4.6\% & 16.0$\pm$4.5\% & 30.4$\pm$5.7\% & 45.2$\pm$6.1\% & \textbf{56.4$\pm$5.7\%} \\
\midrule
\textbf{Average} & 8.4$\pm$3.6\% & 18.6$\pm$4.1\% & 26.8$\pm$4.9\% & 38.1$\pm$5.4\% & \textbf{49.0$\pm$4.9\%} \\
\bottomrule
\end{tabular*}
\vspace{-8pt}
\end{table*}

\begin{figure*}[t!]
  \centering
\includegraphics[width=\textwidth]{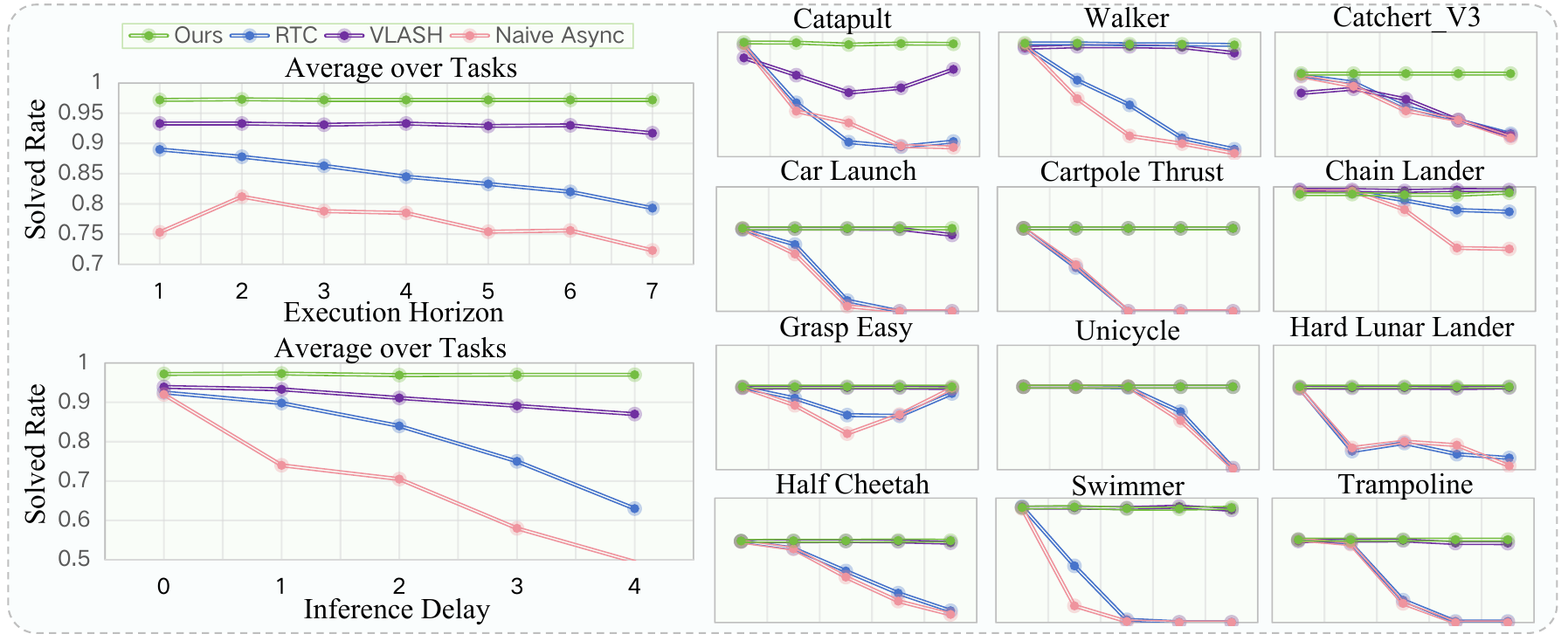}
\vspace{-10pt}
  \caption{Performance comparison on the Kinetix benchmark. \textbf{Top left:} Execution horizon vs. average solve rate with a fixed inference delay of 1. \textbf{Bottom left:} Inference delay vs. average solve rate with a fixed execution horizon = max (inference delay, 1). \textbf{Right:} Inference delay vs. solve rate for individual tasks. Each setting consists of 2048 trials.}
  \label{fig3}
  \vspace{-10pt}
\end{figure*}

\section{Experiments}
\label{sec:experiments}


\subsection{Simulation Benchmark on Kinetix}

\paragraph{Setup and Baselines.}
We utilize Kinetix \cite{b37}, a benchmark specifically designed for evaluating dynamic policies. To ensure a fair comparison in this state-based environment, we adapt our Flow-Planner to an MLP-based policy predicting action chunks ($H=8$), while the Res-Refiner serves as a lightweight residual network. Note that the optical flow module is omitted here as Kinetix provides direct state observations. We benchmark against two SOTA real-time manipulation methods: RTC \cite{b24} and VLASH \cite{b25} (Vision-based DynamicVLA \cite{b38} cannot work in state-based Kinetix simulation benchmark). System latency is simulated by introducing an artificial delay $\delta \in \{0, 1, 2, 3, 4\}$ steps between observation and execution.

\paragraph{Overall Performance.}
Figure \ref{fig3} illustrates the aggregate success rate across 12 Kinetix tasks. Under the minimal latency ($\delta=0$), all methods perform comparably. However, as the inference delay increases---mimicking the computation load of large VLA models---the performance of RTC and VLASH degrades rapidly. In contrast, DSDyn-VLA exhibits exceptional robustness, maintaining a success rate of over 97\% even at the most severe latency setting ($\delta=4$). 

\begin{table*}[t!]
\centering
\footnotesize
\vspace{-15pt}
\caption{Ablation study of average success rates on Kinetix under varying latencies ($\delta$) and execution horizons ($K$). We evaluate different execution sizes for each delay setting. DSDyn-VLA demonstrates superior performance compared to ablated variants.}
\vspace{-5pt}
\label{tab:kinetix_ablation_delay_exec}
\newcolumntype{Y}{>{\centering\arraybackslash}X}
\begin{tabularx}{\textwidth}{l *{8}{Y} c}
\toprule
\multirow{2}{*}{} & \multicolumn{2}{c}{Delay $\delta=0$} & \multicolumn{2}{c}{Delay $\delta=1$} & \multicolumn{2}{c}{Delay $\delta=2$} & \multicolumn{2}{c}{Delay $\delta=3$} & \multicolumn{1}{c}{Delay $\delta=4$} \\
\cmidrule(lr){2-3} \cmidrule(lr){4-5} \cmidrule(lr){6-7} \cmidrule(lr){8-9} \cmidrule(lr){10-10}
Execution Horizon (K)& $1$ & $8$ & $1$ & $7$ & $2$ & $6$ & $3$ & $5$ & $4$ \\
\midrule
\rowcolor[gray]{0.9} DSDyn-VLA & 97.2\% & 97.2\% & 97.3\% & 97.1\% & 96.9\% & 97.3\% & 97.0\% & 97.3\% & 97.1\% \\
\textit{w/o Future} & 97.0\% & 97.1\% & 93.6\% & 93.1\% & 91.6\% & 91.9\% & 88.8\% & 89.6\% & 85.6\% \\
\textit{w/o Refiner} & 93.9\% & 90.9\% & 93.3\% & 91.7\% & 91.1\% & 91.3\% & 89.1\% & 89.1\% & 87.0\% \\
\bottomrule
\end{tabularx}
\vspace{-5pt}
\end{table*}


\begin{table}[t]
    \centering
    \vspace{-5pt}
    \footnotesize
    \setlength{\tabcolsep}{4pt}

    \caption{Results on our DynBench. Each method is evaluated with 100 trials for each task.}
    \vspace{-12pt}
    \label{tab:dynbench}
    \begin{tabular}{lcccccccccc}
        \toprule
        Method & Pick & Drop & Stack & Pour & Sort & Insert & Ball & Can & GreenBall & Avg \\
        \midrule
        PI0.5                  & 12\% & 14\% &  1\% & 10\% & 19\% &  1\% & 18\% & 10\% &  2\% &  9.7\% \\
        RTC                    & 16\% & 41\% &  7\% & 13\% & 25\% &  1\% & 32\% & 25\% & 15\% & 19.4\% \\
        VLASH                  & 19\% & 55\% &  8\% & 30\% & 37\% & 12\% & 35\% & 32\% & 16\% & 27.1\% \\
        DynamicVLA             & 28\% & 55\% & 19\% & 48\% & 43\% & 17\% & 44\% & 36\% & 19\% & 34.3\% \\
        \midrule
        DSDyn-VLA w/o Refiner  & 49\% & 56\% & 32\% & 54\% & 49\% & 19\% & 46\% & 48\% & 31\% & 42.7\% \\
        DSDyn-VLA w/o Future   & 37\% & 58\% & 30\% & 54\% & 53\% & 24\% & 50\% & 42\% & 36\% & 42.7\% \\
        DSDyn-VLA w/o Flow     & 36\% & 48\% & 23\% & 50\% & 33\% & 17\% & 44\% & 37\% & 30\% & 35.3\% \\
        \rowcolor[gray]{0.9}
        DSDyn-VLA              & \textbf{51\%} & \textbf{62\%} & \textbf{37\%} & \textbf{64\%} & \textbf{53\%} & \textbf{25\%} & \textbf{53\%} & \textbf{51\%} & \textbf{40\%} & \textbf{48.4\%} \\
        \bottomrule
    \end{tabular}
    \vspace{-15pt}
\end{table}

\begin{wrapfigure}[13]{r}{0.5\columnwidth}
    \vspace{-20pt}
    \centering
    \includegraphics[width=\linewidth]{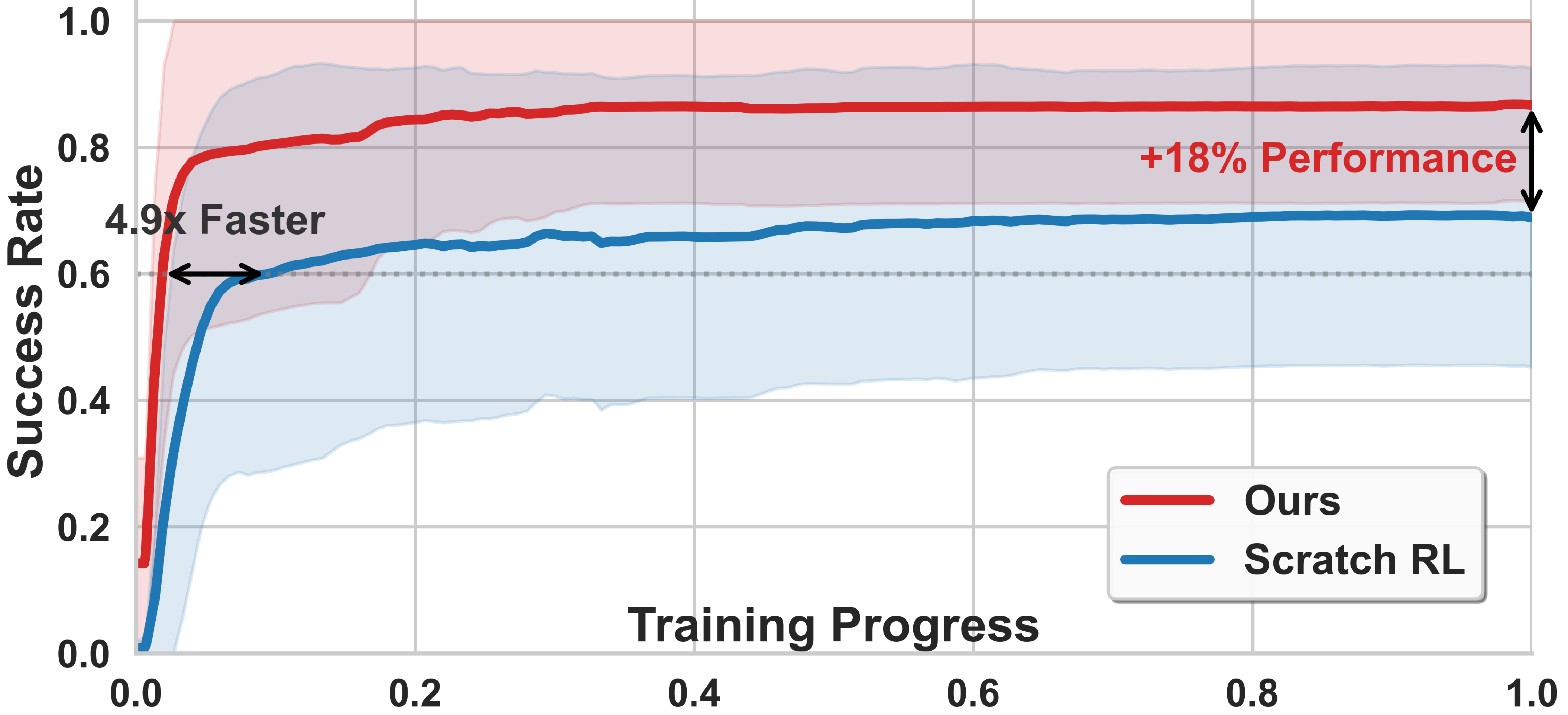}
    
    \caption{Training efficiency and performance comparison on Kinetix tasks. The \textbf{Res-Refiner} (Ours, red) significantly outperforms the standard RL baseline (Blue), achieving \textbf{4.9x faster convergence} and a \textbf{+18\% higher success rate}.}
    \label{fig:efficiency}
\end{wrapfigure}

\paragraph{Impact of Modules.}
The quantitative results in Table \ref{tab:kinetix_ablation_delay_exec} isolate the contribution of each component. Removing the future prediction mechanism (\textit{w/o Future}) leads to a significant performance drop ($-11.5\%$ at $\delta=4$). Furthermore, removing the fast stream (\textit{w/o Refiner}) results in unstable execution ($-10.1\%$), highlighting that the open-loop execution of chunks fails to correct for errors, whereas the Res-Refiner successfully stabilizes the trajectory.

\paragraph{Training Efficiency and Residual Efficacy.}
To verify that the performance gains stem from the Dual-Stream framework rather than simply training a strong RL policy, we conducted an analysis on 12 Kinetix tasks. Figure \ref{fig:efficiency} compares the training curves of our \textit{Res-Refiner} against a standard RL policy trained from scratch with the same network architecture. Results indicate that the Res-Refiner demonstrates significantly superior sample efficiency, achieving convergence \textbf{4.9x faster} than Scratch-RL. Moreover, the average peak success rate of Res-Refiner is substantially higher, delivering a \textbf{+18\% performance gain} over the baseline. This suggests that learning a \textit{residual} on top of Flow-planner is a much easier optimization problem than learning global control dynamics from scratch, validating the ``Slow-Fast'' design philosophy.

\subsection{DynBench Experiments}
\paragraph{Details of DynBench.} DynBench is a MuJoCo\cite{b39}-based benchmark for dynamic manipulation, designed to evaluate the full visual pipeline of DSDyn-VLA as a complement to the state-based Kinetix benchmark. We construct DynBench by porting four real-world dynamic tasks into simulation with a Franka Panda robot arm and further expanding them into a 9-task suite, including Conveyor Picking, Dynamic Dropping, Dynamic Stacking, Mobile Pouring, category-aware conveyor sorting, cylinder insertion into a moving hole, grasping a rolling ball, grasping a rolling can, and selecting the green ball when red and green balls roll in together. Each task is accompanied by 100 rule-generated demonstrations to support consistent training and evaluation.

\begin{figure*}[t!]
\vspace{-15pt}
  \centering
  \includegraphics[width=0.9\textwidth]{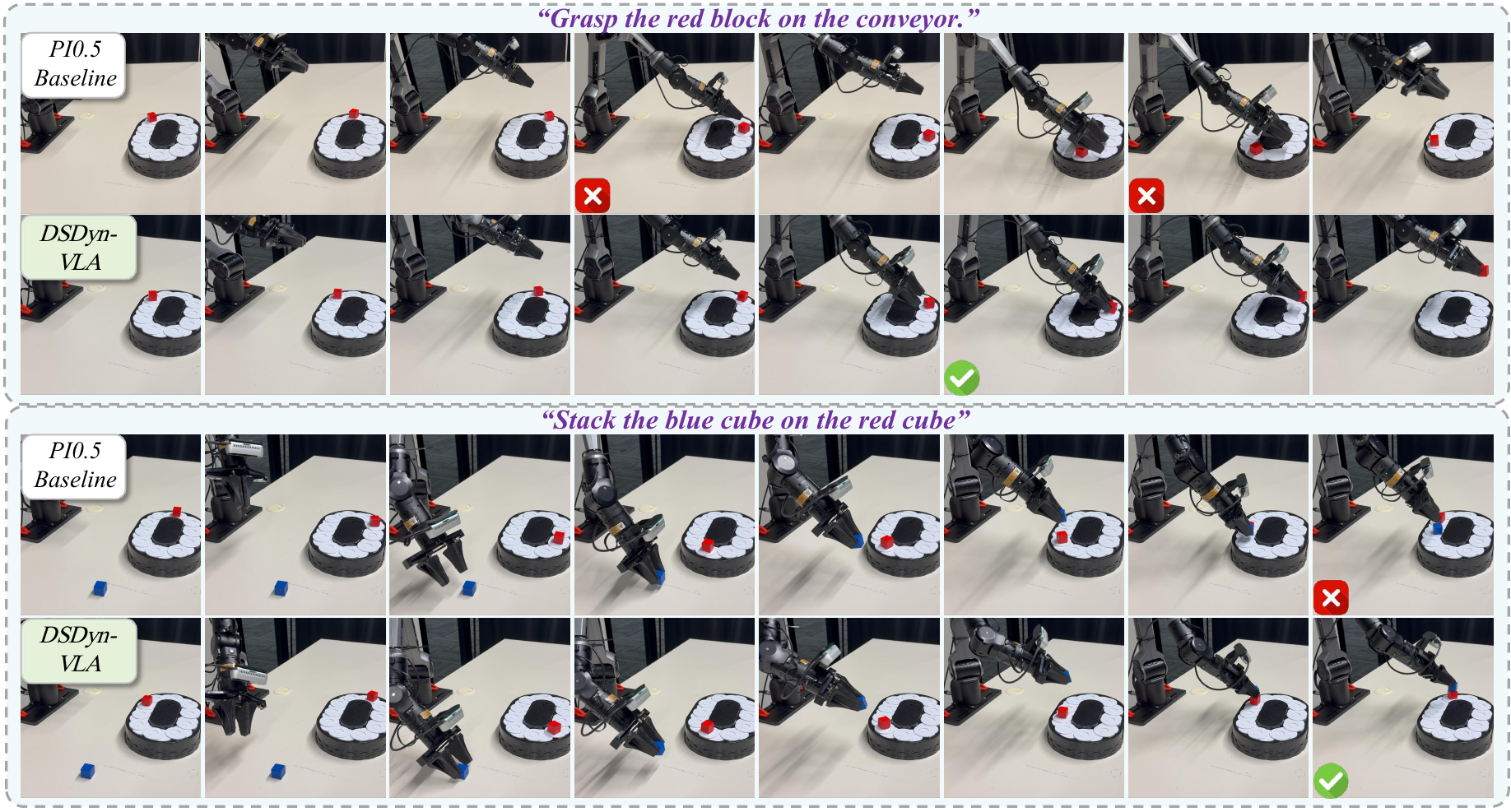}
  \vspace{-10pt}
  \caption{Case comparison of our DSDyn-VLA and baseline (PI0.5).}
  \label{fig5}
  \vspace{-15pt}
\end{figure*}

\paragraph{Results on DynBench.}
We further evaluate both baseline methods and ablated variants on DynBench, with 100 trials for each task. As shown in Table~\ref{tab:dynbench}, DSDyn-VLA achieves the best average success rate of 48.4\%, substantially outperforming PI0.5, RTC, VLASH, and DynamicVLA. The ablation results further verify the effectiveness of all three key components: removing the Refiner or Future module reduces the average success rate to 42.7\%, while removing the Flow branch leads to a larger drop to 35.3\%. These results show that Dyn-Bench provides a controlled yet visually grounded testbed, and further confirms the advantage of our full system under dynamic manipulation settings.

\subsection{Real-World Experiments}

\paragraph{Hardware and Task Setup.}
We validate DSDyn-VLA on a real-world robotic setup using a 6-DOF Agilex Piper arm equipped with two Intel RealSense D435 cameras. The Slow Stream is powered by a \textbf{PI0.5} backbone, fine-tuned with a chunk size of $H=10$. The Res-Refiner operates at 30Hz. We design four challenging dynamic tasks: (1) \textbf{Conveyor Picking}, where the robot must grasp a target cube moving on a conveyor belt; (2) \textbf{Dynamic Dropping}, where the robot must release a held object into a moving container; (3) \textbf{Dynamic Stacking}, requiring the robot to pick a cube and stack it onto another moving cube; and (4) \textbf{Mobile Pouring}, a high-precision task where the robot pours water into a moving cup. In the real-world setting, the binary reward for RL is provided manually: once the task is observed to be successfully completed, a reward of 1 is assigned. RL stage takes about 30-60 mins on each task.


\begin{wraptable}[11]{r}{0.50\columnwidth}
\vspace{-14pt}
\centering
\footnotesize
\caption{Design choice ablation on four real-world task (50 trials). We compare the full model against variants with alternative visual encodings or training strategies.}
\label{tab:real_world_ablation}
\vspace{-6pt}
\begin{tabular}{p{0.32\columnwidth}c}
\toprule
Method Variant & Success Rate \\
\midrule
\textbf{DSDyn-VLA (Full)} & \textbf{48.5\%} \\
\midrule
\textit{w/o Optical Flow} & 30.5\% \\
\textit{w/o Polar Map} & 33.0\% \\
\textit{w/o Future State Awareness} & 36.5\% \\
\textit{w/o Res-Refiner} & 25.0\% \\
Res-Refiner (\textit{w/ SFT}) & 37.5\% \\
\bottomrule
\end{tabular}
\vspace{-10pt}
\end{wraptable}

\paragraph{Performance Analysis.}
Table \ref{tab:real_world_results} quantifies the substantial gap between comparative baselines and our method over 5$\times$50 trials per task. 

\paragraph{Ablation on Design Choices.}
To justify our specific architectural decisions, we performed a fine-grained ablation study on four \textit{Real-World} tasks (Table \ref{tab:real_world_ablation}).

\paragraph{1) Impact of Optical Flow (\textit{w/o Optical Flow}).}
Removing the optical flow input completely causes a catastrophic drop in success rate to 30.5\%. This result confirms that standard static RGB inputs lack the necessary high-order derivative information (velocity) required to track moving objects. Without explicit motion cues, the VLA acts on outdated positional information, failing to intercept the target.


\paragraph{2) Impact of Polar Mapping (\textit{w/o Polar Map}).}
Replacing our polar mapping with a ResNet-34 encoder that projects raw flow directly into the VLA's token space yields a suboptimal success rate of 33.0\%. We attribute this to \textit{feature space misalignment}: the projected tokens from a scratch-trained encoder lack the semantic grounding expected by the VLA backbone. In contrast, our polar mapping translates motion into color and utilize CLIP as encoder, effectively aligning geometric dynamics with the VLA's inherent semantic understanding of the color space.

\paragraph{3) Impact of Future State Awareness (\textit{w/o Future}).}
Disabling the Future State Awareness mechanism results in a success rate of 36.5\%. Without this module, the Flow-Planner generates actions relative to the robot's state at the moment of observation ($t$), ignoring the system latency ($T_{lat}$). By the time these actions are executed at $t+T_{lat}$, the robot has moved, leading to a coordinate frame mismatch. The drop in performance validates the necessity of conditioning generation on the execution-time state.

\begin{wraptable}[29]{r}{0.49\columnwidth}
\vspace{-30pt}
\centering
\footnotesize
\captionof{table}{Module-level runtime profiling.}
\label{tab:module_profile}
\vspace{-4pt}
\begin{tabular}{p{0.31\columnwidth}c}
\toprule
Component & Latency \\
\midrule
VLA backbone total & 73 ms \\
\quad -image encoder & 14 ms \\
\quad -VLM prefill & 32 ms \\
\quad -action expert decoding & 27 ms \\
System latency & 13 ms \\
RAFT-Large & 3 ms \\
Flow encoder & 14 ms \\
Kinematics rollout & $<1$ ms \\
\midrule
Flow-Planner total & 103 ms \\
Res-Refiner total & 3 ms \\
\bottomrule
\end{tabular}

\vspace{-5pt}

\captionof{table}{Residual magnitude statistics on Dynamic Stacking and Mobile Pouring.}
\label{tab:residual_budget}
\begin{tabular}{p{0.36\columnwidth}c}
\toprule
Statistic & Ratio \\
\midrule
All dims $<80\%$ of range & 95.8\% \\
All dims $<95\%$ of range & 99.1\% \\
\bottomrule
\end{tabular}

\vspace{-5pt}

\caption{Robustness under degraded optical-flow conditions, averaged over four real-world tasks.}
\label{tab:flow_robustness}
\begin{tabular}{p{0.32\columnwidth}c}
\toprule
Condition & Avg. Success \\
\midrule
Clean & 48.5\% \\
20\% occlusion & 47.0\% \\
Low light & 45.5\% \\
High light & 46.0\% \\
Additive noise & 45.5\% \\
\midrule
50\% occlusion & 43.5\% \\
70\% occlusion & 42.0\% \\
Reflective stickers & 42.5\% \\
Blinking light & 40.5\% \\
\bottomrule
\end{tabular}

\vspace{-26pt}
\end{wraptable}

\paragraph{4) Impact of Closed-Loop Correction (\textit{w/o Res-Refiner}).}
Removing the \textit{Res-Refiner} drops performance to 25.0\%, highlighting the severity of the \textit{Control Gap}. Even with perfect planning, open-loop execution of an action chunk ($10$ steps) leaves the robot blind to real-time perturbations. The Res-Refiner's ability to inject high-frequency corrections (30Hz) is critical for adjusting the end-effector pose during the final approach phase.

\paragraph{5) Impact of RL Training (Res-Refiner \textit{w/ SFT}).}
Training the Res-Refiner via SFT instead of RL recovers performance only to 37.5\%, significantly lower than the RL-trained counterpart. We attribute this to the distribution shift problem: SFT policies struggle to recover from states that deviate from expert demonstrations. In contrast, our RL-based refiner actively learns a robust recovery policy, proving that the reinforcement learning formulation is essential for the fast stream's reactivity.

\paragraph{Robustness to degraded optical flow.}
We further evaluate DSDyn-VLA under degraded visual conditions to test its sensitivity to imperfect optical flow. As shown in Table~\ref{tab:flow_robustness}, the performance drops gradually rather than collapsing, indicating that DSDyn-VLA is robust to degraded optical flow.

\paragraph{Module-level runtime profiling.}
We profile each module on a RTX4090. The additional motion-perception and residual-correction components are lightweight compared with the base VLA backbone, introducing only modest runtime overhead, as summarized in Table~\ref{tab:module_profile}.

\paragraph{Is the residual range sufficient?}
The Res-Refiner predicts residuals bounded to 10\% of the full action range per dimension. To verify that this constraint is not overly restrictive, we measure the residual magnitudes on Dynamic Stacking and Mobile Pouring. Table~\ref{tab:residual_budget} shows that the residual branch rarely saturates, suggesting that this budget is sufficient for local closed-loop correction.

\section{Conclusion}
We present DSDyn-VLA, a dual-stream framework for dynamic manipulation that integrates motion perception, future-state awareness, and real-time residual correction into a unified control paradigm. Across both simulation and real-world tasks, DSDyn-VLA consistently outperforms strong baselines. While there remains room to further broaden residual-policy generalization and expand real-world evaluation across more tasks and platforms, the current results already suggest strong potential for applications such as industrial handling and assistive robotics. More broadly, this work may broaden the applicability of robotic manipulation in dynamic settings, while also raising the need for careful handling of safety and failure risks in human-shared environments.

\subsection*{AI use statement}

In this work, we used generative AI tools to edit the manuscript to improve its readability and to assist with formatting references. We have not used generative AI tools for generating synthetic datasets, developing theoretical models or conceptual frameworks, formulating or proving mathematical claims, proposing or refining hypotheses, designing or providing feedback on research methodology or experiments, implementing methods, interpreting results, or cleaning and reformatting datasets. We have reviewed all AI-assisted edits to the text. We take responsibility for the final content of this work, including text, claims, or artifacts produced with the aid of generative AI.

\subsection*{Reproducibility statement}

We provide source code for DSDyn-VLA and DynBench, including the training and evaluation pipelines for the Flow-Planner and Res-Refiner, together with demonstration videos, in the supplementary materials. Section~3 describes the model architecture and two-stage training procedure. Section~4 presents the experimental setups for Kinetix, DynBench, and real-world tasks, while Appendix~A.1 reports additional evaluations on LIBERO and DOMINO. Model weights will be released upon publication.

\bibliography{iclr2027_conference}
\bibliographystyle{iclr2027_conference}

\appendix
\section{Appendix}
\subsection{External Benchmark Experiments}

\paragraph{Static Manipulation on LIBERO.}
Since the Res-Refiner does not take text embeddings as input, it is necessary to verify that the added streams of DSDyn-VLA do not hurt performance on general static manipulation tasks that require precise instruction following. We evaluate DSDyn-VLA on the four LIBERO \cite{b19} suites: LIBERO-Spatial, LIBERO-Object, LIBERO-Goal, and LIBERO-10. As shown in Table~\ref{tab:libero}, DSDyn-VLA matches or outperforms the PI0.5 baseline on all four suites, indicating that the introduced motion perception and residual correction streams preserve the static manipulation and instruction-following capability of the underlying VLA. We attribute this to the fact that the Res-Refiner learns via RL to repair suboptimal actions, which also reduces failures on static tasks.

\paragraph{Dynamic Manipulation on DOMINO.}
To further validate generalization beyond Kinetix and DynBench, we evaluate DSDyn-VLA on DOMINO \cite{b49}, a recently proposed benchmark featuring complex object dynamics for dynamic manipulation. We compare against the PI0.5 baseline and PUMA, the state-of-the-art dynamics-aware method introduced in the same work, using the success rate (SR) and the manipulation score (MS), a continuous execution-quality metric. As shown in Table~\ref{tab:domino}, DSDyn-VLA surpasses PUMA on both metrics, improving SR from 17.18\% to 20.74\% and MS from 35.08 to 41.64. These results on an external dynamic benchmark further confirm the effectiveness and generalizability of DSDyn-VLA.

\section{Related Work}
\label{sec:appendix_related}
\subsection{RL Finetuning for VLA Models}

RL has proven critical for enhancing VLA robustness by exploring corner cases uncovered by SFT. While methods like VLAC \cite{b30} and VLA-RL \cite{b31} have successfully fine-tuned autoregressive VLAs, they are incompatible with Flow Matching architectures. Although some theoretical works explore RL for flow-based policies \cite{b32,b33}, they are difficult to scale. PI0.6* \cite{b12} recently demonstrated real-world fine-tuning of a flow-based VLA. However, it relies on a binarized advantage, which reduces RL to a weakened form of conditional behavior cloning. Our DSDyn-VLA addresses this by training a external non-flow module via online RL to augment the VLA.

\subsection{Methods for Perception Gap}
For the perception gap, prior work on dynamic manipulation has mainly explored two directions. One line \cite{b40,b41,b42} uses event cameras for fast visual servoing and grasping, benefiting from high temporal resolution and low latency, but these methods typically rely on specialized neuromorphic sensors and a more limited data ecosystem than standard RGB pipelines. Another line \cite{b43,b44} exploits optical flow for dynamic control and policy learning, showing that motion cues are valuable for manipulation; however, such methods are often task-specific and do not naturally interface with large RGB-pretrained VLA backbones. In contrast, our method stays within a standard RGB setup and introduces motion cues through a polar-mapped optical-flow representation, which is more compatible with CLIP-style visual encoders.

\begin{table}[t]
    \centering
    \small
    \setlength{\tabcolsep}{4.5pt}
    \caption{Success rates (\%) on the four LIBERO suites (mean $\pm$ std over 3 seeds). The dynamic-oriented components of DSDyn-VLA do not degrade performance on static instruction-following tasks.}
    \label{tab:libero}
    \begin{tabular}{lcccc}
        \toprule
        Method & Spatial & Object & Goal & LIBERO-10 \\
        \midrule
        PI0.5     & 98.67$\pm$0.23 & 98.30$\pm$0.10 & 97.87$\pm$0.42 & 92.60$\pm$0.17 \\
        \rowcolor[gray]{0.9} DSDyn-VLA & \textbf{99.17$\pm$0.15} & \textbf{98.73$\pm$0.06} & \textbf{98.17$\pm$0.29} & \textbf{93.30$\pm$0.30} \\
        \bottomrule
    \end{tabular}
    \vspace{-8pt}
\end{table}

\begin{table}[t]
    \centering
    \small
    \caption{Results on the DOMINO benchmark for dynamic manipulation (mean $\pm$ std over 3 seeds).}
    \label{tab:domino}
    \begin{tabular}{lcc}
        \toprule
        Method & SR (\%) & MS \\
        \midrule
        PI0.5     & 9.63$\pm$0.08  & 26.19$\pm$0.41 \\
        PUMA      & 17.18$\pm$0.16 & 35.08$\pm$0.13 \\
        \rowcolor[gray]{0.9} DSDyn-VLA & \textbf{20.74$\pm$0.25} & \textbf{41.64$\pm$0.66} \\
        \bottomrule
    \end{tabular}
    \vspace{-8pt}
\end{table}

\subsection{Methods for Control Gap}
For the control gap, prior embodied-control methods have improved execution-time reactivity mainly through continual replanning or closed-loop action updating. Representative examples include video-prediction \cite{b45} or MPC-style replanning \cite{b46}, receding-horizon control \cite{b47}, and test-time resampling for chunked actions \cite{b48}, all of which aim to restore sensitivity to the latest observation during execution. These methods highlight the importance of closing the loop online, but often require repeated full replanning or expensive sampling. By contrast, our approach preserves the VLA's chunk-level planning ability while introducing a lightweight high-frequency residual correction module, making it more suitable for dynamic manipulation under nontrivial inference latency.

\section{Detailed Ablation Results on Kinetix Benchmark}
We provide the detailed results of the ablation experiments on Kinetix in Table~\ref{tab:kinetix_ablation_tasks}.
\begin{table*}[t!]
\centering
\small
\caption{Detailed ablation results on 12 Kinetix tasks under varying latencies ($\delta \in \{0, 2, 4\}$) and execution horizon of 4. We compare the full DSDyn-VLA against variants lacking future state awareness (\textit{w/o Future}) or the residual refinement stream (\textit{w/o Refiner}).}
\vspace{-5pt}
\label{tab:kinetix_ablation_tasks}
\setlength{\aboverulesep}{0pt}
\setlength{\belowrulesep}{0pt}
\renewcommand{\arraystretch}{1.15} 
\setlength{\tabcolsep}{0pt}
\begin{tabular*}{\textwidth}{
    @{\extracolsep{\fill}} 
    c  l cccccc 
    @{\extracolsep{0pt}} 
    !{\hspace{15pt}} 
    >{\columncolor{gray!15}}c !{\color{gray!15}\vrule width 12pt} 
    >{\columncolor{gray!15}}c !{\color{gray!15}\vrule width 12pt} 
    >{\columncolor{gray!15}}c
}
\toprule
\multirow{2}{*}{Type} & \multirow{2}{*}{Task} & \multicolumn{3}{c}{\textit{w/o Future}} & \multicolumn{3}{c}{\textit{w/o Refiner}} & \multicolumn{3}{c}{\cellcolor{gray!15}DSDyn-VLA} \\
\cmidrule(lr){3-5} \cmidrule(lr){6-8} \cmidrule(lr){9-11}
& & $\delta=0$ & $\delta=2$ & $\delta=4$ & $\delta=0$ & $\delta=2$ & $\delta=4$ & $\delta=0$ & $\delta=2$ & $\delta=4$ \\
\midrule
\multirow{3}{*}{\newblock High-Dyn \newblock} 
  & Catapult & 91.2\% & 49.3\% & 41.5\% & 73.1\% & 51.7\% & 56.1\% & 90.9\% & 92.2\% & 90.6\% \\
  & Walker & 90.9\% & 75.4\% & 0.4\% & 87.4\% & 86.8\% & 86.9\% & 91.0\% & 89.9\% & 89.7\% \\
  & Catcher\_V3 & 99.9\% & 99.8\% & 99.7\% & 73.5\% & 70.4\% & 21.7\% & 99.7\% & 99.7\% & 99.7\% \\
\midrule
\multirow{9}{*}{Regular} 
  & Car Launch & 99.5\% & 99.2\% & 99.5\% & 98.9\% & 99.4\% & 99.3\% & 99.4\% & 99.5\% & 99.4\% \\
  & Cartpole Thrust & 100\% & 100\% & 100\% & 100\% & 100\% & 100\% & 100\% & 100\% & 99.9\% \\
  & Chain lander & 95.2\% & 95.9\% & 97.5\% & 97.6\% & 97.9\% & 97.5\% & 94.6\% & 95.2\% & 95.3\% \\
  & Grasp Easy & 99.9\% & 100\% & 100\% & 99.0\% & 99.0\% & 99.0\% & 99.9\% & 99.8\% & 100\% \\
  & Unicycle & 100\% & 100\% & 100\% & 100\% & 100\% & 100\% & 100\% & 100\% & 100\% \\
  & Hard Lunar Lander & 100\% & 100\% & 99.8\% & 98.2\% & 98.8\% & 98.6\% & 100\% & 100\% & 100\% \\
  & Half Cheetah & 98.7\% & 98.0\% & 97.8\% & 98.5\% & 98.0\% & 97.9\% & 98.9\% & 97.9\% & 98.7\% \\
  & Swimmer & 92.2\% & 92.6\% & 91.3\% & 92.9\% & 92.2\% & 93.4\% & 92.2\% & 92.5\% & 92.2\% \\
  & Trampoline & 99.5\% & 99.5\% & 99.9\% & 98.1\% & 99.2\% & 94.1\% & 99.7\% & 99.8\% & 99.3\% \\
\midrule
\multicolumn{2}{l}{\textbf{Avg (12 Tasks)}} & 97.2\% & 92.5\% & 85.6\% & 93.1\% & 91.1\% & 87.0\% & 97.2\% & 97.2\% & 97.1\% \\
\bottomrule
\end{tabular*}
\vspace{-8pt}
\end{table*}

\section{Fairer Temporal Baselines and Optical-Flow Encoding Variants}
\label{sec:fair_baselines_flow_variants}

To provide a fairer comparison with prior dynamic-manipulation baselines, we strengthen \textsc{RTC}, \textsc{VLASH}, and \textsc{DynamicVLA} with two additional historical frames and further perform $1$ hour of end-to-end PPO training for each method. This setting reduces the concern that our gains might simply come from providing richer temporal context or additional policy adaptation.

In addition, we conduct several variants on our full model \textsc{DSDyn-VLA} to better understand the role of motion representation. Specifically, we compare the full model against: (1) replacing optical flow with simple frame stacking, (2) directly encoding optical flow with a ResNet-based encoder, and (3) encoding optical flow with a learnable MLP adapter. The results are reported in Table~\ref{tab:fair_baselines_flow_variants}.

We make three observations. First, adding historical frames and PPO training consistently improves all strengthened baselines: \textsc{RTC} improves from $18.0$ to $24.0$, \textsc{VLASH} improves from $26.5$ to $32.5$, and \textsc{DynamicVLA} improves from $37.5$ to $39.0$. This confirms that temporal context and additional policy optimization are indeed helpful for dynamic manipulation. However, even after these enhancements, all baselines remain clearly below \textsc{DSDyn-VLA}, which achieves $48.5$ average success.

Second, replacing optical flow with naive frame stacking leads to a noticeable drop from $48.5$ to $42.5$. This suggests that the benefit of our method does not come merely from exposing the model to more temporal observations; rather, explicitly modeling motion is more effective than asking the network to infer it implicitly from stacked RGB frames.

Third, alternative optical-flow encoding strategies are substantially less effective than our full design. Direct optical-flow encoding yields only $33.0$, while a learnable MLP adapter reaches $37.0$, both significantly below the full \textsc{DSDyn-VLA}. These results indicate that the performance gain is not obtained by simply adding an extra flow branch. Instead, the way motion cues are represented and aligned with the visual backbone is critical for effective dynamic manipulation.

Overall, these results strengthen two conclusions: (i) our advantage cannot be explained solely by stronger temporal input or extra RL fine-tuning, and (ii) explicit and properly encoded motion representation is a key ingredient behind the performance of \textsc{DSDyn-VLA}.

\begin{table}[t]
\centering
\small
\caption{Fairer comparison with temporally strengthened baselines and ablations on motion representation. For a fair comparison, we augment \textsc{RTC}, \textsc{VLASH}, and \textsc{DynamicVLA} with two historical frames and $1$ hour of end-to-end PPO training. We further compare several motion-input variants of our \textsc{DSDyn-VLA}.}
\label{tab:fair_baselines_flow_variants}
\resizebox{0.6\linewidth}{!}{
\begin{tabular}{lc}
\toprule
\textbf{Setting} & \textbf{Avg.\ Success (\%)} \\
\midrule
RTC & 18.0 \\
RTC + history frames + PPO & 24.0 \\
VLASH & 26.5 \\
VLASH + history frames + PPO & 32.5 \\
DynamicVLA & 37.5 \\
DynamicVLA + history frames + PPO & 39.0 \\
\midrule
DSDyn-VLA & \textbf{48.5} \\
DSDyn-VLA w/ frame stacking & 42.5 \\
DSDyn-VLA direct encoding & 33.0 \\
DSDyn-VLA learnable MLP & 37.0 \\
\bottomrule
\end{tabular}
  \vspace{-10pt}
}
\end{table}

\begin{figure*}[t!]
  \centering
  \includegraphics[width=\textwidth]{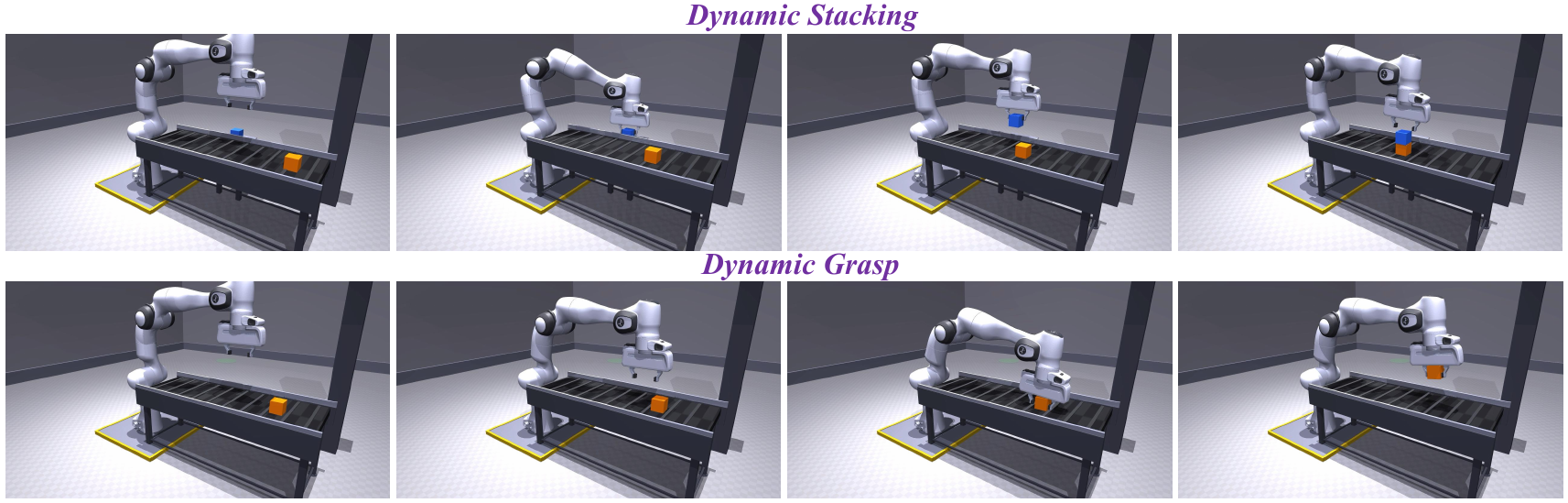}
  \vspace{-10pt}
  \caption{Qualitative rollouts of DSDyn-VLA on DynBench. Top: Dynamic Stacking. Bottom: Dynamic Grasp. From left to right, the snapshots show that DSDyn-VLA tracks moving objects and completes the task under continuous motion.}
  \label{fig6}
  \vspace{-15pt}
\end{figure*}

\end{document}